\documentclass[runningheads]{llncs}
\usepackage{arxiv}

\usepackage{graphicx}
\usepackage{booktabs}

\usepackage[accsupp]{axessibility} 
\usepackage[pagebackref,breaklinks,colorlinks,citecolor=blue]{hyperref}

\usepackage{orcidlink}

\usepackage{cite}
\usepackage{amsmath,amssymb,amsfonts}
\usepackage{caption}
\usepackage{graphicx}
\usepackage{textcomp}
\usepackage{algorithm}
\usepackage{algpseudocode}
\usepackage{comment}
\usepackage{tabularx}
\usepackage{multirow}
\usepackage[font=small,labelfont=bf]{caption}

\begin{document}

\title{event-ECC: Asynchronous Tracking of Events with Continuous Optimization}


\author{Maria Zafeiri \inst{1} 
\orcidlink{0009-0007-7820-6649}\\ 
\And Georgios Evangelidis \inst{2} 
\orcidlink{1111-2222-3333-4444}\\ 
\And Emmanouil Psarakis \inst{1} 
\orcidlink{2222--3333-4444-5555}
\and
\\
\inst{1} Department of Computer Engineering and Informatics, University of Patras
, Greece 
\\
\texttt{\{zafeiri, psarakis\}@ceid.upatras.gr} \\ \and
\inst{2}Snap Inc., Vienna, Austria 
\\ 
\email{georgios@snap.com}
}

\maketitle

\begin{abstract}
In this paper, an event-based tracker is presented. 
Inspired by recent advances in asynchronous processing of individual events, we develop a direct matching scheme that aligns spatial distributions of events at different times. 
More specifically, we adopt the Enhanced Correlation Coefficient (ECC) criterion and propose a tracking algorithm that computes a 2D motion warp per single event, called event-ECC (eECC). 
The complete tracking of a feature along time is cast as a \emph{single} iterative continuous optimization problem, whereby every single iteration is executed per event. 
The computational burden of event-wise processing is alleviated through a lightweight version that benefits from incremental processing and updating scheme.
We test the proposed algorithm on publicly available datasets and we report improvements in tracking accuracy and feature age over state-of-the-art event-based asynchronous trackers.
\end{abstract}

\section{Introduction}
\label{sec:intro}
Visual tracking constitutes a core component in odometry~\cite{nister_CVPR2005_visual_odometry} and SLAM~\cite{davison_PAMI07_monoslam} pipelines. Commonly, such solutions rely on frames from conventional cameras, thus making feature tracking in high-speed or high dynamic range (HDR) scenarios very challenging. Rather, event-based tracking appears to be more robust in these conditions, mainly because event cameras capture small brightness changes, irregularly and asynchronously, with microsecond latency~\cite{gallego_PAMI20_eventVisionSurvey}.
 
Event-based tracking has its own challenges though. One of the main difficulties is the establishment of correspondences between events, because of the varying scene appearance and its dependency on camera or scene motion. 
Early event-based tracking methods adopted frame-based formulations and shaped regular frames from event streams to process them synchronously, optionally along with intensity images. 
More recent algorithms that process only events in an asynchronous manner have been also gaining attention ~\cite{alzugaray_3DV19_haste,alzugaray_BMVC20_haste}. 
However, they usually rely on discrete optimization schemes, thus decreasing the tracking accuracy, in order to keep the complexity low, and in turn the compromised tracking duration (a.k.a. feature age).

In this paper, we adopt the ECC-based image alignment formulation~\cite{evangelidis_PAMI08_ECC, psarakis2005enhanced} to introduce a novel event tracking algorithm that process events individually. Inspired by the approach of recurring 2D density maps from event streams~\cite{alzugaray_3DV19_haste}, we propose a Gauss-Newton optimization with \textit{an update step per event}, exploiting the asynchronous nature of event cameras. 
Unlike the conventional optimization-based tracking~\cite{gehrig_ECCV18_featureTracking} that solves an iterative  optimization problem per observation (frame or event-set), {the proposed scheme solves \textit{a single optimization problem} per complete track}, whereby an incremental optimization step (one iteration) updates the tracking state per event.

The contributions of this work are summarized as follows:
\begin{itemize}
    \item We adjust an image alignment optimization-based scheme to an asynchronous event-based tracking algorithm that estimates a non-discrete motion warp per event, by a single optimization step. Thereby, the complete tracking of a single feature is formulated as a single continuous optimization problem.
    
    \item A lightweight equivalent version that benefits from event-wise incremental processing is proposed.
        
    \item In the context of asynchronous event tracking, the proposed algorithm attains state-of-the-art performance in terms of accuracy and feature lifetime.
\end{itemize}
\section{Related Work}
\label{sec:related_work}
In order to establish correspondences in time, event-based trackers build event descriptors (features) either from single conventional frames \cite{gehrig_IJCV20_EKLT, tedaldi_EBCCSP16_featureDetectionDavis} or by integrating events temporally \cite{zhu_ICRA17_probabilisticDataAssociation, alzugaray_3DV19_haste, alzugaray_BMVC20_haste}. 
Since the events are generated more frequently around the image's edges, they are typically combined with frames and intensity gradients to build reference patches. 
\cite{tedaldi_EBCCSP16_featureDetectionDavis} and \cite{kueng_IROS16_lowLatencyVO} build point-set templates from Canny edge maps which  are then matched with event batches using the Iterative Closed Point (ICP) optimization~\cite{besl_PAMI1992_ICP}. Similarly, \cite{gehrig_IJCV20_EKLT} aligns intensity gradient templates with event-based brightness images using ECC optimization~\cite{evangelidis_PAMI08_ECC}. More interestingly, when the event-sensing is solely used, such templates are built from raw events. In this context, \cite{zhu_ICRA17_probabilisticDataAssociation} proposes a probabilistic formulation to align motion-compensated events with newly generated events; this tracker is integrated into a Visual Inertial Odometry (VIO) solution in~\cite{zhu_CVPR17_visualInertialOdometry}. Likewise, \cite{rebecq_BMVC17_keyframe} generates motion-corrected event frames and aligns them using Lucas-Kanade tracker~\cite{lucas_IJCAI1981_iterativeImageRegistration} within a VIO pipeline. 

The aforementioned methods have high complexity and might not fully leverage the asynchronous nature of data. Event-driven feature detectors~\cite{vasco_IROS16_harrisCorner, li_IROS19_FAHarris, Manderscheid_CVPR19_cornerDetectorDeepLearning} favor the asynchronous processing but the tracking of such features may not be of sufficient quality~\cite{mueggler_BMVC17_cornerDetection, alzugaray_RoboticsAutomationLetters18_CornerDetection}. To build reference patches that can be tracked for longer time, \cite{alzugaray_3DV19_haste} projects motion-compensated events into a 2D density map (template) that is refined in an event-by-event fashion. Such a template is compared against an instant density map (model), built from the recent raw events, over a discretized space of solutions. A significant speedup of this tracker is achieved by \cite{alzugaray_BMVC20_haste} through some approximations that enable incremental processing per event. Recent clustering~\cite{hu_IROS22_eCDT} and learning-based~\cite{messikomer_CVPR23_featureTracking} trackers obtain longer feature age than {those proposed in}~\cite{alzugaray_3DV19_haste}. 

The proposed method adopts the formulation of \cite{alzugaray_3DV19_haste, alzugaray_BMVC20_haste} to build template and model images from a fixed-size circular buffer of events. However, unlike searching over multiple state update proposals, an incremental ECC-based~\cite{evangelidis_PAMI08_ECC} algorithm aligns the template and model instances {in an event-wise based way}. That is, a single optimization step (iteration) is executed per event. 

\section{Problem Formulation}
\subsection{Preliminaries}
Let us denote by $\mathcal{S}_{I}$ the image support area of 2D integer coordinates $\mathbf{n}\in\mathbb{Z}^2$, and by $\mathcal{X}_{\mathcal{S}_{I}}$ the corresponding area with 2D real coordinates 
$\mathbf{x}~\in~\mathbb{R}^2$.

Let us also consider that for any chosen feature $F$, its state  at time instance $T$ is denoted by the following $3\times1$ vector: 
\begin{equation}\label{eq:state}
    \mathbf{s}=\left [
         \mathbf{x}_F(T)^{t} ~
         \theta_{F}(T)
    \right]^t
\end{equation}
\noindent with $\mathbf{x}_F(T)$ being the 2D position of feature $F$ in $\mathcal{X}_{\mathcal{S}_{I}}$ at that time instance,  and $ \theta_{F}(T
)$ the orientation of its neighborhood $\mathcal{N}\Big(\mathbf{s}(T)\Big)$ of size $(2N+1)\times (2N+1)$, defined by:
\begin{equation}
\mathcal{N}\big(\mathbf{s}\big)=\Big\{  \mathbf{x}\in~\mathcal{X}_{\mathcal{S}_{I}}~ \big|~||\mathbf{x}-\mathbf{x}_{F}(T)||_2\leq N~\Big\}\label{neighbor}
\end{equation}
\noindent with respect to the global coordinate system 
and $||\mathbf{z}||_2$ denoting the $l_2$ norm of vector $\mathbf{z}$. It is clear that 
$\mathcal{N}\big(\mathbf{s}_0\big)$ is of special form, derived from Eq. (\ref{neighbor})
with $\mathbf{x}_{F}(0)\in \mathcal{S}_I$ and $\theta_{F}(0)=0^\circ$. We use this set to define the support of the associated template patch, that is:
\begin{equation}
\mathcal{S}_{\mathcal{T}_F}=\Big\{ \mathbf{n}'=\mathbf{n}-\mathbf{x}_F(0),~\forall~\mathbf{n}\in~\mathcal{N}\big(\mathbf{s}_0\big)\Big\}\label{T_support},
\end{equation}
while we define the densities of this area, as well as the ones of the model window, in the next subsection.

Let us also denote by $\mathbf{e}=\left [T~\mathbf{x}(T)^t \right]^t$ the event generated at the position $\mathbf{x}$ of $\mathcal{X}_{\mathcal{S}_{I}}$, at time $T$.\footnote{We silently assume here rectified coordinates after undistortion.} 

If we now assume a Euclidean geometric transformation parameterized by {the elements of the state vector defined in Eq. (\ref{eq:state})}, that is, for each pair of corresponding points $\mathbf{x}(T)~\in~\mathcal{N}\big(\mathbf{s}\big)$ and $\mathbf{x}'(T)~\in~\mathcal{X}_{\mathcal{S}_{\mathcal{T}_{F}}}$, then $
\mathbf{x}(T)=R\big(\theta_F(T)\big)\mathbf{x}'(T)+\mathbf{x}_{F}(T)$,
or equivalently:
\begin{equation}
\mathbf{x}'(T)=R^T\big(\theta_F(T)\big)\big(\mathbf{x}(T)-\mathbf{x}_{F}(T)\big)\label{tem_in1}
\end{equation}
\noindent with $\mathcal{X}_{\mathcal{S}_{\mathcal{T}_{F}}}$ being the continuous counterpart of $\mathcal{S}_{\mathcal{T}_F}$ in Eq. (\ref{T_support}). 

Finally, let us define the four-pixel neighborhood associated to an event generated at $\mathbf{x}$, as
\begin{equation}
    \mathcal{N} = \Big\{ \mathbf{n},~ \mathbf{n} + \mathbf{v}_1,~ \mathbf{n} + \mathbf{v}_2,~ \mathbf{n} + \mathbf{v}_1 +\mathbf{v}_2  \Big\}\label{set_N},
\end{equation}
\noindent where the vectors $\mathbf{v}_i,~~i=1,2$ constitute the natural basis of $\mathbb{R}^2$ and $ \mathbf{n}= \bigl\lfloor \mathbf{x} \bigr \rfloor$ denotes the element-wise floor of $\mathbf{x}$. Then, similar to~\cite{alzugaray_3DV19_haste}, we define a density  update rule using linear interpolation as follows: 
\begin{equation}
\mathcal{I}(\mathbf{n},T_k)\longleftarrow \mathcal{I}(\mathbf{n},T_{k-1})+
\prod_{i=1}^2\big((2\Delta x_{i}-1)n_i+(1-\Delta x_i)\big)\label{int_scheme}
\end{equation}
\noindent for each $\mathbf{n}\in \mathcal{N}$, with $\Delta \mathbf{x}=\mathbf{x}(T_k)-\mathbf{n}$.

\subsection{Initialization Phase}\label{subsec_init_phase}
Let us consider a circular buffer of $2M+1$ events, indexed by the time of the first event, 
\begin{equation}
    \mathcal{E}_{\mathcal{M}}(k)=\big\{  \mathbf{e}_m\big\}_{m=0}^{2M}\label{events_set}
\end{equation}
\noindent with  $ \mathcal{E}_{\mathcal{M}}(0)$ being the set of the first $2M+1$ detected events. Similar to~\cite{alzugaray_BMVC20_haste}, we use the $ \mathcal{E}_{\mathcal{M}}(0)$ to initialize the model and the template windows,  $\mathcal{M}_F(\mathbf{n},0)$ and $\mathcal{T}_{F}(\mathbf{n}',0),~\mathbf{n}'\in~\mathcal{S}_{\mathcal{T}_{F}}$, respectively. Given the initial state and the initial event-set, we use the interpolation scheme (Eq. (\ref{int_scheme})) for each event $\mathbf{e}_m$ to define the initial template and model:
\begin{equation}   \mathcal{T}_F(\mathbf{n}_m',0)=\mathcal{M}_F(\mathbf{n}_m,0)\equiv\mathcal{I}(\mathbf{n}_m,0)
\end{equation}
\noindent with $\mathbf{n}_m'\in\mathcal{S}_{\mathcal{T}_F}$. 
For instance, the active (non-zero) area of $\mathcal{M}_F(\mathbf{n},0)$ is defined by the union of all the 4-point neighborhoods of every event $\mathbf{e}_m$, namely $\bigcup_{m=0}^{2M}\mathcal{N}_m$. Likewise, the active area of the template is defined by the respective neighborhoods of the transformed events' positions according to the initial state and Eq. (\ref{tem_in1}). Finally, we define the vectorized forms of these two patches, as $\mathbf{m}_F(0)$ and $\mathbf{t}_F(0)$ respectively.

\subsection{Tracking the Feature $F$}
The feature's tracking starts when the $2M+1$-th event is detected. 
Let us now consider that at time $T_{k}\geq~0$:
the state vector $\mathbf{s}_{k}$ and the content of set $\mathcal{E}_{\mathcal{M}}(k)$ are known, and that at time $T_{k+1}$ the $(k+1+2M)$-th event is detected.
{Then the following two steps should be done:} 
\begin{itemize}
 \item{$\mathbf{S}_1$}: the set $\mathcal{E}_{\mathcal{M}}(k)$ should be updated as follows:
\begin{equation}
    \mathcal{E}_{\mathcal{M}}(k+1)=\mathcal{E}_{\mathcal{M}}(k)\backslash \big\{~\mathbf{e}_{k}~\big\}~\cup~\big\{~\mathbf{e}_{k+1+2M}~\big\},\label{update_event_set}
    \end{equation}
    where $``\backslash"$ is the set minus operator with $\mathcal{A}\backslash \mathcal{B} =\{\mathbf{x}: \mathbf{x}~\in~\mathcal{A},~\text{and}~\mathbf{x}~\notin~\mathcal{B}\}$ and $\cup$ the set union operator 
\item{$\mathbf{S}_2$}: the density map of the model $\mathcal{M}_F(\mathbf{n},~T_{k+1})$ should be computed and those pixels of the template density map $\mathcal{T}_{F}(\mathbf{n}',~T_{k+1})$ which correspond to the central event of $\mathcal{E}_{\mathcal{M}}(k+1)$, located at $\mathbf{x}_c(T_{k+1})$, should be updated.
\end{itemize}

While $\mathbf{S}_1$ is very straightforward, $\mathbf{S}_2$ requires the knowledge of the state vector $\mathbf{s}_{k+1}$ in order to apply the motion compensation according to Eq. (\ref{tem_in1}), that is:
\begin{equation}
\mathbf{x}_c'(T_{k+1})=R^T\big(\theta_F(T_{k+1})\big)\big(\mathbf{x}_c(T_{k+1})-\mathbf{x}_F(T_{k+1})\big)\label{poin_x}
\end{equation}
with $\mathbf{x}_c'(T_{k+1})\in\mathcal{X}_{\mathcal{T}_F}$ and $\mathcal{X}_{\mathcal{T}_F}$ denoting the continuous version of the template's support area $\mathcal{S}_{\mathcal{T}_F}$.
Then, we can properly use Eq. (\ref{set_N}) to define the quantities $\mathcal{N}'_{k+1}$, $\mathbf{n}'_{k+1} = \bigl\lfloor \mathbf{x}_c'(T_{k+1}) \bigr \rfloor $ and the residuals $ \Delta \mathbf{x}'=\mathbf{x}_c'(T_{k+1})-\mathbf{n}'_{k+1}$, and use the interpolation scheme (Eq. (\ref{int_scheme})) to update the appropriate template's density map $\mathcal{T}_{F}(\mathbf{n}',T_{k+1}),~\forall~\mathbf{n}'~\in~\mathcal{N}'_{k+1}$. 

In the next subsection, we are going to estimate the state vector $\mathbf{s}_{k+1}$ which is required for the update step $\mathbf{S}_2$. 

\section{Event tracking using ECC Criterion}\label{sec:proposed_method}

\subsection{Model and Template Windows}
After updating the set $\mathcal{E}_{\mathcal{M}}(k+1)$ 
and using Eq. (\ref{set_N}), we locate the corresponding neighborhood  $\mathcal{N}_m$ per event $\mathbf{e}_m$, and use the interpolation scheme (Eq. (\ref{int_scheme})) to calculate the density map of the model window $\mathcal{M}_F(\mathbf{n}_m,~T_{k+1})$, 
thus forming the model vector $\mathbf{m}_F(T_{k+1})$ of length $(2N+1)^2$. 

Having formed the model window, we use the motion model of Eq. (\ref{tem_in1}) to warp every non-zero density $\mathcal{M}_F(\mathbf{n}_m,~T_{k+1})$ of each pixel $\pmb{n}_{m}$ into the position:
\begin{equation}
\mathbf{x}'_m\big(\mathbf{s}_{k+1}\big)=R^{T}\big(\theta_F(T_{k+1})\big)\big(\pmb{n}_{m}-\pmb{x}_F(T_{k+1})\big)\label{xc'}
\end{equation}
of the support $\mathcal{X}_{\mathcal{S}_{\mathcal{T}_F}}$ of the template, that is:  
\begin{equation}\mathcal{T}_F\Big(\mathbf{x}'_m\big(\mathbf{s}_{k+1}\big),~T_{k+1}\Big)=\mathcal{M}_F(\mathbf{n}_m,~T_{k+1})=\mathcal{T}_F\big(\mathbf{n}_m',~\mathbf{s}_{k+1}\big).\nonumber
\end{equation}
Note that those elements of the warped template vector  $\mathbf{t}_F\big(\mathbf{s}_{k+1}\big)$ 
which correspond to model's pixel with zero density, attain their previous values $\mathcal{T}_F\big(\mathbf{n}_m'~T_{k}\big)$. 

\subsection{The Proposed Optimization Criterion}
We then propose the use of the following well-known ECC criterion \cite{psarakis2005enhanced, evangelidis_PAMI08_ECC} to quantify the performance of the motion warp with parameters $\mathbf{s}_{k+1}$:
\begin{equation}
C_{ECC}\big(\mathbf{s}_{k+1}\big)=\Bigg|\Bigg|\frac{\mathbf{t}_F\big(\mathbf{s}_{k+1}\big)}{||\mathbf{t}_F\big(\mathbf{s}_{k+1}\big)||_2}-\frac{\mathbf{m}_F(T_{k+1})}{||\mathbf{m}_F(T_{k+1})||_2}\Bigg|\Bigg|_2^2
\label{ECCCr}\end{equation}
and its minimization w.r.t. the parameters of the state vector, that is:
\begin{equation}  
\mathbf{s}^{\star}_{k+1}=\underset{\mathbf{s}_{k+1}}{\text{argmin}} ~~C_{ECC}\big(\mathbf{s}_{k+1}\big).\label{optimization}
\end{equation}
Solving the optimization problem is clearly not a simple task because of the nonlinearity involved in the correspondence part.
This, of course, suggests that its minimization requires nonlinear optimization techniques by adopting a gradient-based approach. 
To this end, let us consider the following additive updating rule for each one of the elements of the state vector:
\begin{equation}
\mathbf{s}_{k+1}=\mathbf{s}_{k}+\Delta\mathbf{s}_{k+1}\label{Ds}
\end{equation}
where
$
\mathbf{s}_{k}=\left[\mathbf{x}_F(T_{k})^T~\theta_F(T_{k})\right]^T$ and
$\Delta\mathbf{s}_{k+1}=
\left[\Delta\mathbf{x}_F(T_{k+1})^T~\Delta\theta_F(T_{k+1})\right]^T.$
By substituting Eq. (\ref{Ds}) into Eq. (\ref{xc'}) and assuming  that $\Delta\theta_F (T_{k+1})$ takes on small values, after some mathematical manipulations, we obtain: 
\begin{equation}
     \pmb{x}_m'\big(\mathbf{s}_{k+1}\big) ~ \approx \pmb{x}'_{m}\big(\mathbf{s}_{k}\big)+W_m\big(\mathbf{s}_{k}\big)\Delta\pmb{s}_{k+1}
\label{linear}\nonumber
\end{equation}
where $\pmb{x}_m'\big(\mathbf{s}_{k+1}\big)~\in~\mathcal{X}_{\mathcal{S}_{\mathcal{T}_F}}$ and the $2\times3$ matrix $W_m\big(\mathbf{s}_{k}\big)$ depends only on the state vector $\mathbf{s}_{k}$.
Note that $ \pmb{x}_m'\big(\mathbf{s}_{k+1}\big)$ is composed by two terms. The first one depends on the state vector $\mathbf{s}_{k}$ which is known, while the second one  depends on the unknown perturbation vector $\Delta\pmb{s}_{k+1}$. Therefore, using first order Taylor approximation around $\mathbf{s}_{k}$ for every element of 
the warped vector $\mathbf{t}_F\big(\mathbf{s}_{k+1}\big)$, we obtain:
\begin{equation}
\mathbf{t}_F\big(\mathbf{s}_{k+1}\big)\approx\mathbf{t}_F\big(\mathbf{s}_{k}\big)+J\big(\mathbf{s}_{k}\big)\Delta\pmb{s}_{k+1}\nonumber
\end{equation}
where $J\big(\mathbf{s}_{k}\big)$ denotes the $(2N+1)^2\times3$ Jacobian matrix of the warped intensity vector $\mathbf{t}_F\big(\mathbf{s}_{k}\big)$, evaluated at the nominal parameter values $\mathbf{s}_{k}$, and $\Delta\mathbf{s}_{k+1}$ denotes the perturbations of the parameters of the state vector. By substituting this approximation into Eq. (\ref{ECCCr}) we get the following simpler objective function:
\begin{equation}   \tilde{C}_{ECC}\big(\Delta\mathbf{s}_{k+1}\big)=\Bigg|\Bigg|
\frac{\mathbf{t}_F\big(\mathbf{s}_{k}\big)+J\big(\mathbf{s}_{k}\big)\Delta\pmb{s}_{k+1}}{||\mathbf{t}_F\big(\mathbf{s}_{k}\big)+J\big(\mathbf{s}_{k}\big)\Delta\pmb{s}_{k+1}||_2}-\frac{\mathbf{m}_F(T_{k+1})}{||\mathbf{m}_F(T_{k+1})||_2}\Bigg|\Bigg|_2^2
\label{ECCCrr}
\end{equation}
which is still non-linear but it benefits from a closed-form optimizer \cite{evangelidis_PAMI08_ECC}. 

To this end, let us define the following quantities:
\begin{eqnarray}
C\big(\mathbf{s}_{k}\big)&=&J^t\big(\mathbf{s}_{k}\big)J\big(\mathbf{s}_{k}\big)\nonumber\\
A\big(\mathbf{s}_{k}\big)&=&C\big(\mathbf{s}_{k}\big)^{-1}J^t\big(\mathbf{s}_{k}\big)\nonumber\\
\mathbf{p}_{\mathbf{t}_F}(\mathbf{s}_{k})&=&J^t(\mathbf{s}_{k})\mathbf{t}_F(\mathbf{s}_{k})\nonumber\\
\mathbf{p}_{\hat{\mathbf{m}}_F}(\mathbf{s}_{k},T_{k+1})&=&J^t(\mathbf{s}_{k})\hat{\mathbf{m}}_F(T_{k+1})\nonumber\\
\rho_{k+1}&=&<\hat{\mathbf{t}}_F\big(\mathbf{s}_{k}\big),~\hat{\mathbf{m}}_F(T_{k+1})> \nonumber\\
\hat{\mathbf{m}}_F(T_{k+1})&=& \frac{\mathbf{m}_F(T_{k+1})}{||\mathbf{m}_F(T_{k+1})||_2}\label{equs}
\end{eqnarray}
where $C\big(\mathbf{s}_{k}\big)$ is a $3\times3$ matrix, $A\big(\mathbf{s}_{k}\big)$ is the $3\times(2N+1)$ pseudo inverse of the jacobian matrix $J\big(\mathbf{s}_{k}\big)$,  $\mathbf{p}_{\mathbf{t}_F}(\mathbf{s}_{k})$ is the projection of template onto the jacobian matrix, $\mathbf{p}_{\hat{\mathbf{m}}_F}(\mathbf{s}_{k},T_{k+1})$ is the projection of model onto the  jacobian matrix, $\rho_{k+1}$ is the correlation coefficient and $\hat{\mathbf{m}}_F(T_{k+1})$ the normalized counterpart of the model vector $\mathbf{m}_F(T_{k+1})$. All those quantities are used in the next lemma that provides the desired result. 

\noindent\textbf{Lemma 1}. 
If $||\mathbf{t}_F(\mathbf{s}_{k})||_2^2>\mathbf{p}_{\mathbf{t}_F}(\mathbf{s}_{k})^tC(\mathbf{s}_{k})^{-1}\mathbf{p}_{\mathbf{t}_F}(\mathbf{s}_{k})$, the optimal perturbation $\Delta\mathbf{s}^{\star}_{k+1}$ needed for the minimization  
of the objective function (\ref{ECCCrr}) is given by:
\begin{equation}
\Delta\mathbf{s}^{\star}_{k+1}=A(\mathbf{s}_{k})
\big(\lambda\hat{\mathbf{m}}_F(T_{k+1})-\mathbf{t}_F(\mathbf{s}_{k})\big)
\label{optimal_solution}
\end{equation}
where $\lambda$ is defined by:
\begin{equation}
\lambda=\frac{||\mathbf{t}_F\big(\mathbf{s}_{k}\big)||_2^2-\mathbf{p}_{\mathbf{t}_F}(\mathbf{s}_{k})^tC(\mathbf{s}_{k})^{-1}\mathbf{p}_{\mathbf{t}_F}(\mathbf{s}_{k})}
{||\mathbf{t}_F\big(\mathbf{s}_{k}\big)||_2\rho_{k+1}-\mathbf{p}_{\mathbf{t}_F}(\mathbf{s}_{k})^tC(\mathbf{s}_{k})^{-1}\mathbf{p}_{\hat{\mathbf{m}}_F}(\mathbf{s}_{k},T_{k+1})}\nonumber.
\end{equation}

\noindent\textbf{Proof.} The proof is based on Theorem 1 of \cite{evangelidis_PAMI08_ECC, psarakis2005enhanced}, it is easy and thus omitted. 
\hfill $\Box$

The computational cost of the optimum perturbation vector $\Delta\mathbf{s}^{\star}_{k+1}$ in Eq. (\ref{optimal_solution}) may be high for an event-based tracking scheme. Therefore, we propose a lightweight verstion of ECC tailored to event-based tracking that benefits from incremental computations per event.

\subsection{Lightweight event-ECC}

At every iteration of event-based ECC, the optimal perturbation $\Delta\mathbf{s}_{k+1}$ requires the computation of several quantities, that is, the jacobian matrix $J(\mathbf{s}_k)$ of size $(2N+1)^2\times3$ in order to get $C(\mathbf{s}_k)$,  the matrix $A(\mathbf{s}_k)$ of size $3\times (2N+1)$ and the template and normalized model vectors $\mathbf{t}_F(\mathbf{s}_k)$, $\hat{\mathbf{m}}_F(T_{k+1})$ of length $(2N+1)^2$. However, note that every newly detected event affects only its 4-pixel neighborhood in the template window. As a result, only a low number of columns and rows need to be updated for some matrices and vectors. Based on that observation, we derive an incremental version of ECC tailored to event-based tracking, with quite reduced complexity.

To this end, consider the set $\mathcal{N}'_{k+1}$ with the coordinates of the four template's  pixels whose densities were updated after the calculation of the warp, defined in Eq. (\ref{optimal_solution}). That is, this set contains the recently modified template's pixels whose densities will be used in the next step of the tracking algorithm. Accordingly, we define the sets $\mathcal{N}'_{{k+1}_x}$ $\mathcal{N}'_{{k+1}_y}$ of the template's gradient whose values have changed. 
Depending on the position of the pixels of $\mathcal{N}'_{k+1}$ in the support area $\mathcal{S}_{T}$ of the template, the cardinality of the union $\mathcal{S}=\mathcal{N}'_{{k+1}_x}\bigcup\mathcal{N}'_{{k+1}_y}$ is bounded by $12$, that is $|\mathcal{S}|\leq~12$.
Having defined the set $\mathcal{S}$ we can define its complement w.r.t. the set $\mathcal{S}_{\mathcal{T}_F}$, that is:
\begin{equation}
\mathcal{S}^c=\mathcal{S}_{\mathcal{T}_F}\backslash\mathcal{S}\nonumber\end{equation}                 
and then, the following quantities:
\begin{eqnarray}
\mathbf{t}_F\big(\mathbf{s}_{k+1}\big)&=&\left[\mathbf{t}_{F_{\mathcal{S}^c}}^t\big(\mathbf{s}_{k}\big)~
\mathbf{t}_{F_{\mathcal{S}}}^t\big(\mathbf{s}_{k+1}\big)\right]^t\nonumber\\
J\big(\mathbf{s}_{k+1}\big)&=&\left[J_{\mathcal{S}^c}^t\big(\mathbf{s}_{k}\big)~
J_{\mathcal{S}}^t\big(\mathbf{s}_{k+1}\big)\right]^t\label{18}
\end{eqnarray}
where the vectors $\mathbf{t}_{F_{\mathcal{S}^c}}\big(\mathbf{s}_{k}\big)$, $\mathbf{t}_{F_{\mathcal{S}}}\big(\mathbf{s}_{k+1}\big)$ and the matrices $J_{\mathcal{S}^c}\big(\mathbf{s}_{k}\big)$, $J_{\mathcal{S}}\big(\mathbf{s}_{k+1}\big)$
constitute a rearrangement of the elements of the corresponding quantities defined in Eq. (\ref{equs}). Therefore, one can easily prove the following equations: 
\begin{eqnarray}
||\mathbf{t}_F\big(\mathbf{s}_{k+1}\big)||_2^2&=&||\mathbf{t}_{F_{\mathcal{S}^c}}(\mathbf{s}_{k})||_2^2+||\mathbf{t}_{F_{\mathcal{S}}}(\mathbf{s}_{k+1})||_2^2\nonumber\\
 C\big(\mathbf{s}_{k+1}\big)&=&
C\big(\mathbf{s}_{k}\big)
+J^t_{\mathcal{S}}\big(\mathbf{s}_{k+1}\big)J_{\mathcal{S}}\big(\mathbf{s}_{k+1}\big)-J^t_{\mathcal{S}}\big(\mathbf{s}_{k}\big)J_{\mathcal{S}}\big(\mathbf{s}_{k}\big)\nonumber\\
 \Delta\mathbf{p}_{k+1}&=&J^t_{\mathcal{S}}\big(\mathbf{s}_{k+1}\big)\mathbf{t}_{F_{\mathcal{S}}}\big(\mathbf{s}_{k+1}\big)\nonumber\\
\Delta\mathbf{p}_{k}&=&J^t_{\mathcal{S}}\big(\mathbf{s}_{k}\big)\mathbf{t}_{F_{\mathcal{S}}}\big(\mathbf{s}_{k}\big)\nonumber\\
 \mathbf{p}_{\mathbf{t}_{F_{\mathcal{S}}}}(\mathbf{s}_{k+1})&=&\mathbf{p}_{\mathbf{t}_{F_{\mathcal{S}}}}(\mathbf{s}_{k})+\Delta\mathbf{p}_{k+1}-\Delta\mathbf{p}_{k}\nonumber\\
 \mathbf{p}_{\mathbf{t}_F}(\mathbf{s}_{k+1})&=&\left[ \mathbf{p}_{\mathbf{t}_{F_{\mathcal{S}^c}}}^t(\mathbf{s}_{k}) ~~\mathbf{p}_{\mathbf{t}_{F_{\mathcal{S}}}}^t(\mathbf{s}_{k+1})\right]^t
 \label{inc_ECC}
\end{eqnarray}
and use them for a quite more efficient computation of the optimal state update in Eq. (\ref{optimal_solution}).

We refer to this algorithm as event-ECC (eECC); the outline is shown in Algorithm \ref{tracking_alg}. Note that, unlike an image alignment scenario where ECC would typically require several iterations, the minimal incremental information here justifies a single iteration per event. 

\begin{algorithm} [ht]

\caption{The Proposed eECC Tracking Algorithm} \label{tracking_alg}
\begin{algorithmic}[eECC]
\State \textbf{Input:} Template vector $\mathbf{t}_{F}(0)$, Neighborhood $\mathcal{N}_{F}\big(\mathbf{x}_F(0)\big)$ of model, Set $\mathcal{E}_{\mathcal{M}}(0)$ and Initial Feature state vector        $\pmb{s}_0$
\Statex \textbf{Output:} Tracking states $S_F=\big[\pmb{s}_0~\pmb{s}_1~\pmb{s}_2\cdots\pmb{s}_L\big]$
\State \textbf{Set} $S_F=\big[\pmb{s}_0\big ]$, $k=0$, $T_{k}=0$

\For {each detected event $\mathbf{e}_m$}
\If{$\mathbf{x}_m~ \in \mathcal{N}_{F}\Big(\mathbf{x}_F(T_{k})\Big)$}
\State \textbf{Update} the set $\mathcal{E}_{\mathcal{M}}(k+1)$ using Eq. (\ref{update_event_set})
\State \textbf{Form} the model's vector $\mathbf{m}_F(T_{k+1})$ using Eqs. (\ref{set_N},\ref{int_scheme}) and  the updated set $\mathcal{E}_{\mathcal{M}}(T_{k+1})$
\State Using Eq. (\ref{optimal_solution}), \textbf{Compute} the optimal  warp $\Delta\mathbf{s}^{\star}_{k+1}$ 
\State \textbf{Update} the state vector $\pmb{s}_{k+1}$ by using Eq. (\ref{Ds})
\State Use $\pmb{s}_{k+1}$ in Eq. (\ref{poin_x}) to find out where the central event $\mathbf{x}_c'(T_{k+1})$ of set $\mathcal{E}_{\mathcal{M}}(k+1)$ is mapped
\State Use the interpolation scheme (Eq. (\ref{int_scheme})) to \textbf{Update} the densities of template's pixels $\in \mathcal{N}'_{k+1}$
\State \textbf{Form} the vector $\mathbf{t}_{F}(\mathbf{s}_{k+1})$ and the jacobian $J(\mathbf{s}_{k+1})$ according to Eq. (\ref{18}) 
\State \textbf{Update} the quantities of Eq. (\ref{inc_ECC})
\State \textbf{Update} the tracking states  $S_F=[S_F~\pmb{s}_{k+1}]$
\State \textbf{Update}  $\mathcal{N}_{F}\Big(\mathbf{x}_F(T_{k+1})\Big)$ according to Eq. (\ref{neighbor})
\State \textbf{Set} $k=k+1$, $T_{k}=T_{k+1}$
\EndIf
\EndFor
\end{algorithmic}
\end{algorithm}

\begin{figure}[ht]
\begin{center}
\begin{minipage}{.425\textwidth} 
   { \begin{tabular}{c c}
   \\
    \includegraphics[width = 0.48\textwidth]{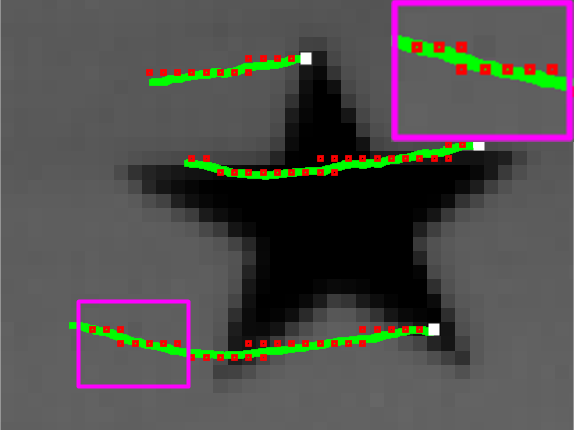}  & 
    \includegraphics[width = 0.48\textwidth]{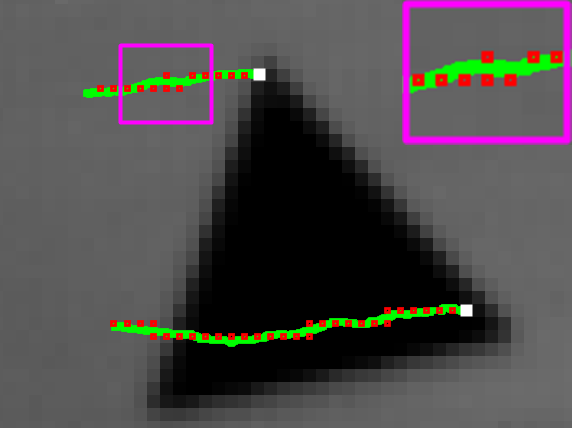}  \\
    \includegraphics[width = 0.48\textwidth]{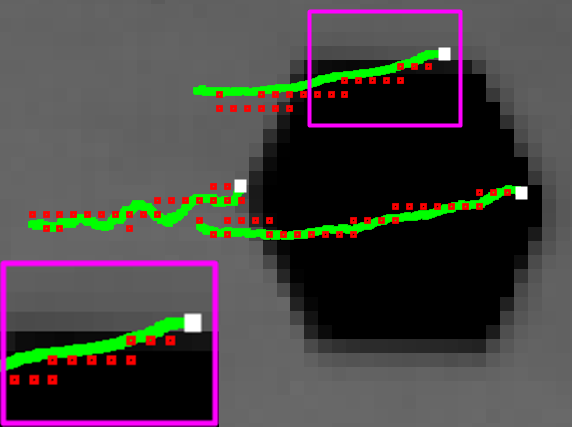}  &
    \includegraphics[width = 0.48\textwidth]{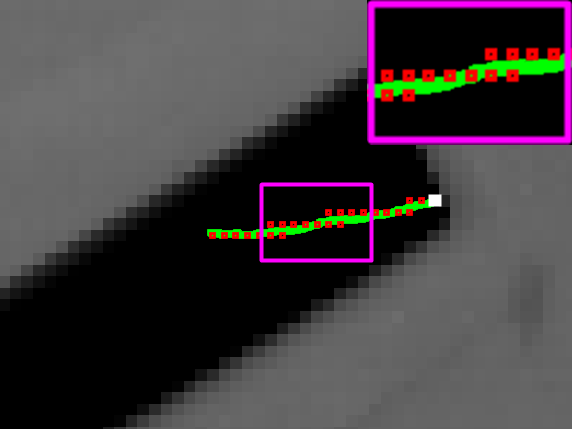} \\
   \\
    \end{tabular} 
    \text{red: Haste Correlation$^\ast$, green: eECC} \\
    }\\   \centering  \hspace{35pt} (a)
    \end{minipage}
    \hspace{50pt}
    \begin{minipage}{.425\textwidth} 
   {     \begin{tabular}{c c}
  \text{Haste Correlation$^\ast$} & \text{eECC} \\
  \includegraphics[width = 0.47\textwidth]{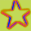}  & 
  \includegraphics[width = 0.47\textwidth]{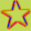} \\
       \includegraphics[width = 0.47\textwidth]{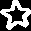}  & 
        \includegraphics[width = 0.47\textwidth]{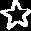} \\
        \text{Active Area: $39\%$} & \text{Active Area: $33\%$} \\
    \end{tabular} }   \centering \hspace{75pt} (b)
    \end{minipage}
    \caption{(a): Tracking trajectories starting from different seeds (white points) in "shapes" scene with 6DoF motion. Unlike HasteCorrelation$^\ast$, eECC generates smooth and continuous trajectories. (b): Final template from tracking the same feature for same age. eECC generates sharper templates (e.g. horizontal edges) because of more accurate tracking and motion compensation. To quantify this difference, we apply the same low threhold into the templates and measure the percentage of the area that generates events. Ideally, such binary masks should resemble the edge-map of the top-left image of Fig.~\ref{fig:mesh1} and this percentage should be less than $10\%$.}
    \label{fig:mesh1}
    \end{center}
\end{figure}

\section{Experiments} \label{sec:experiments}

In this section we test the proposed method on publicly available datasets and compare it against the methods proposed in \cite{alzugaray_3DV19_haste, alzugaray_BMVC20_haste},  \textit{HasteCorrelation$^\ast$} and \textit{HasteDifference$^\ast$} which code is available. These methods achieve state-of-the-art results in asynchronous event-based tracking.
We use the Event Camera dataset \cite{mueggler_IJRR17_dataset} captured by DAVIS240C sensor\cite{brandli_SolidStateCircuits14_visionSensor} that includes APS frames and events at $240\times180$ resolution as well as synced ground truth camera poses at 200Hz from an external motion capture system. To initiate tracking features, we randomly select 500 detected points, uniformly distributed in time, obtained from standard detectors on frames (SURF, FAST, ORB and MSER). All the algorithms are fed with the same set of seeds. Note that frames are not used during tracking and the use of several detectors offers a quite diverse set of features to track.
For a fair comparison, we consider a template window of size $31\times ~31$ and a circular buffer of $193$ events, as used in \cite{alzugaray_BMVC20_haste}. 

The tracking performance is evaluated in terms of two metrics, the tracking accuracy and the feature age, that is, the ability of algorithms to keep the tracks alive. The tracking accuracy is typically quantified by the reprojection error. To reconstruct tracks from multiple views, we use the triangulation method of \cite{nousias_CVPR19_triangulation}. Since the frequency of tracker states is irregular and significantly higher than camera, we select the temporally closest to groundtruth views and then estimate the feature location at these views through linear interpolation. The feature age is estimated through the cumulative percentage of outlies per maximum track lifetime. As in \cite{alzugaray_BMVC20_haste}, we consider a track as outlier when the reprojection error is above $5$ pixels.

\begin{figure}
    \centering
    \begin{tabular}{c c c}
     poster translation & poster rotation & poster 6dof\\
  \includegraphics[width = 0.30\textwidth]{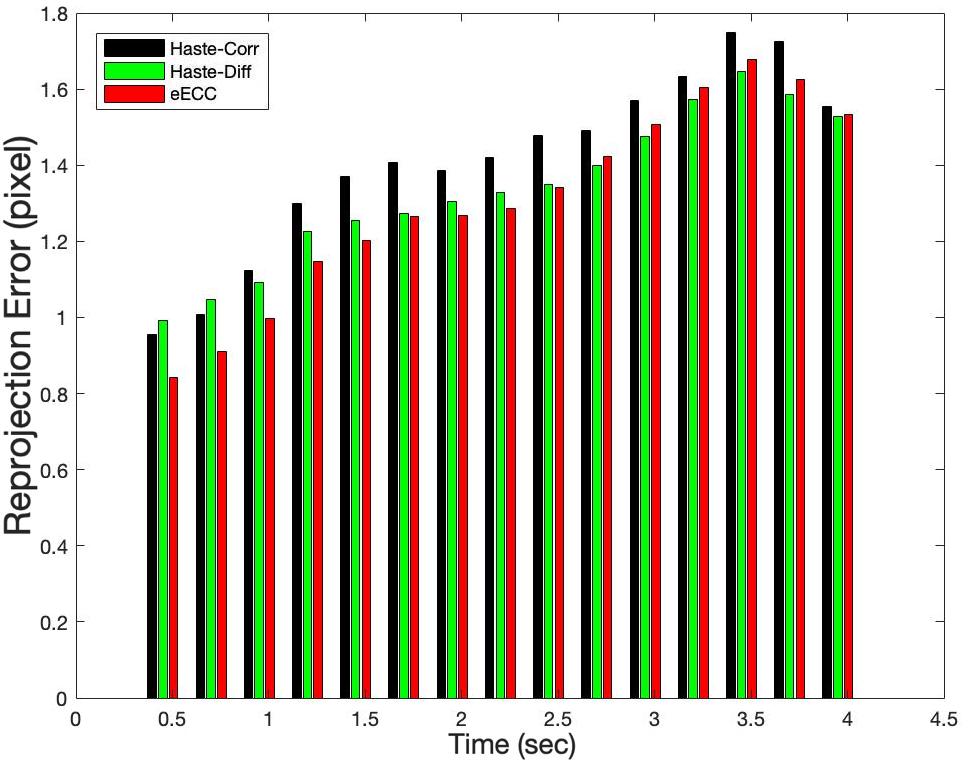}  & 
   \includegraphics[width = 0.30\textwidth]{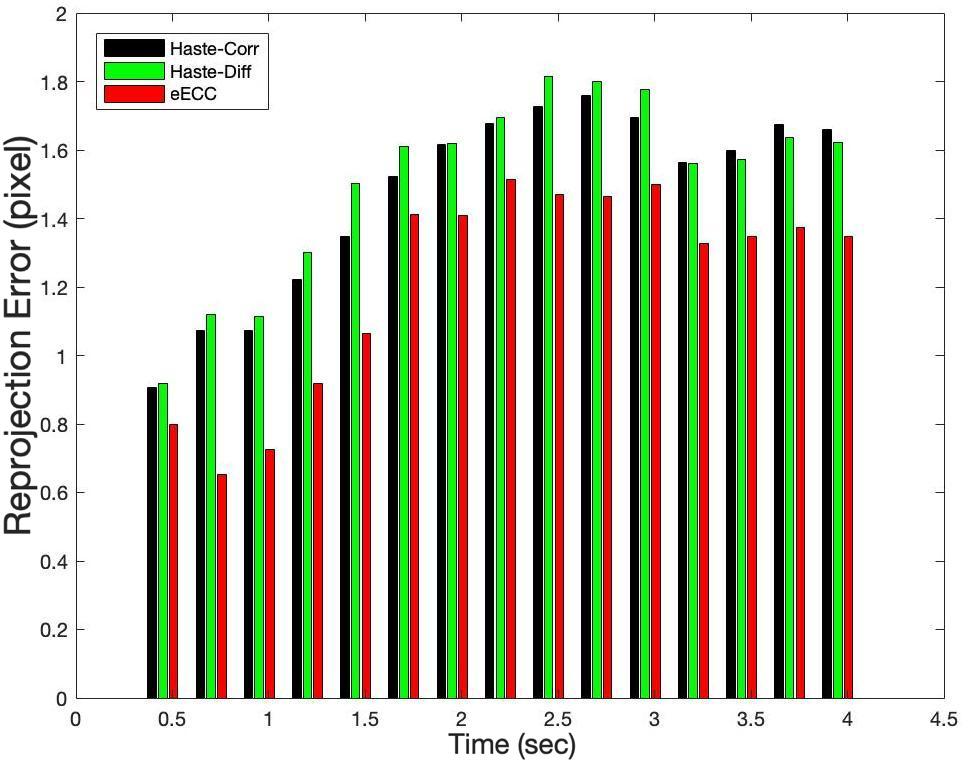}  &
     \includegraphics[width = 0.30\textwidth]{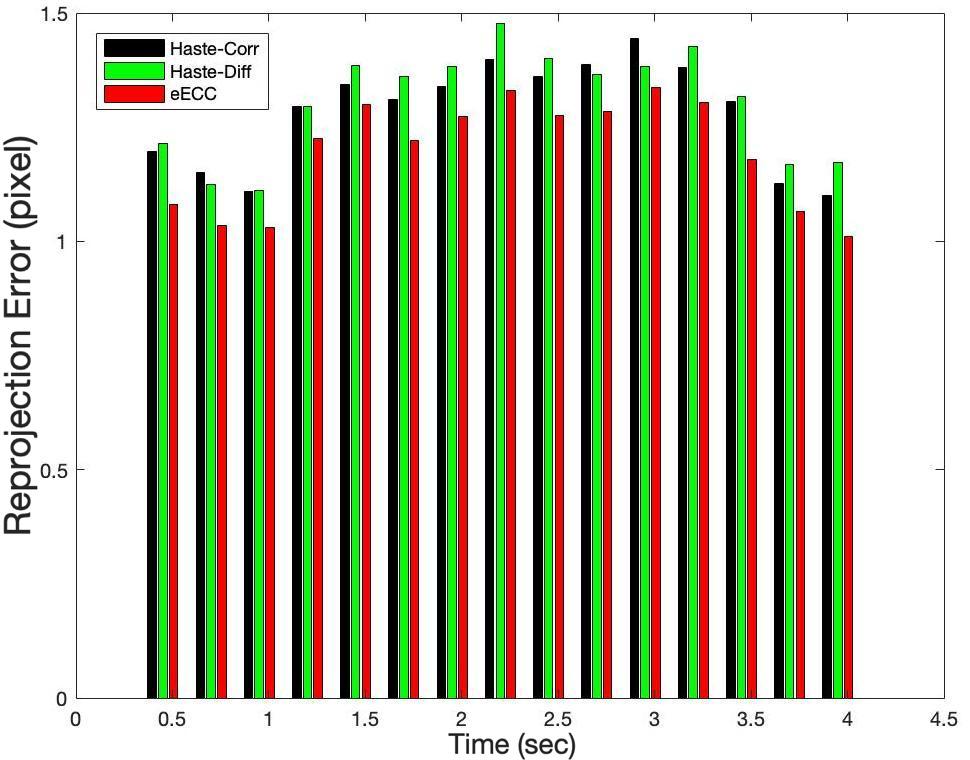} \\
  \includegraphics[width = 0.30\textwidth]{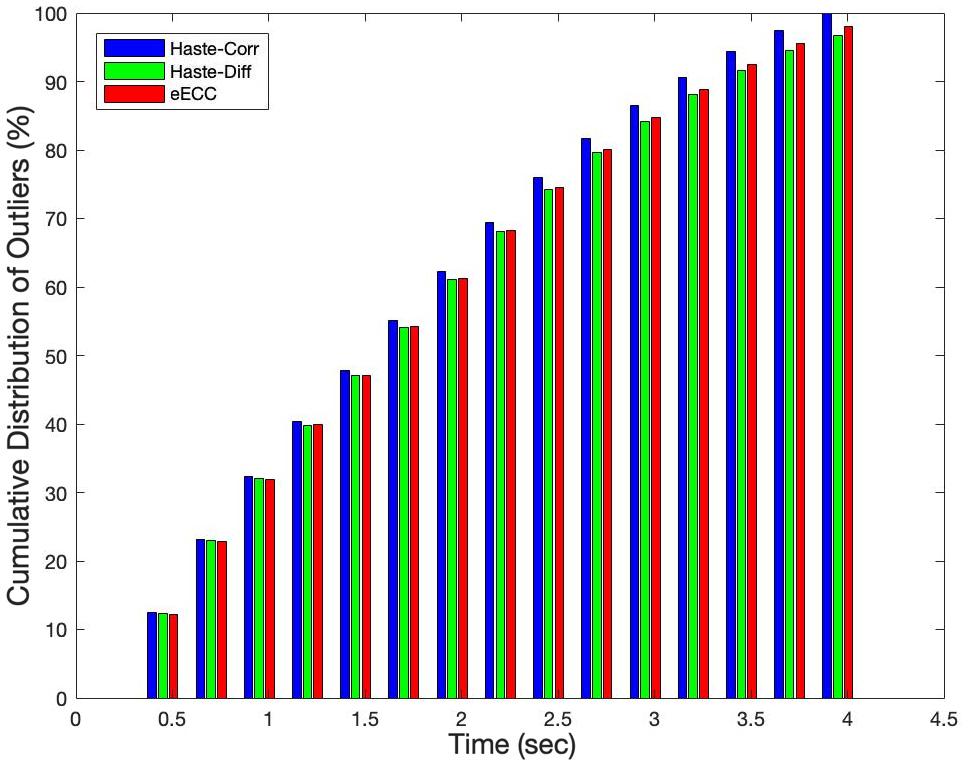}  &  
  \includegraphics[width = 0.30\textwidth]{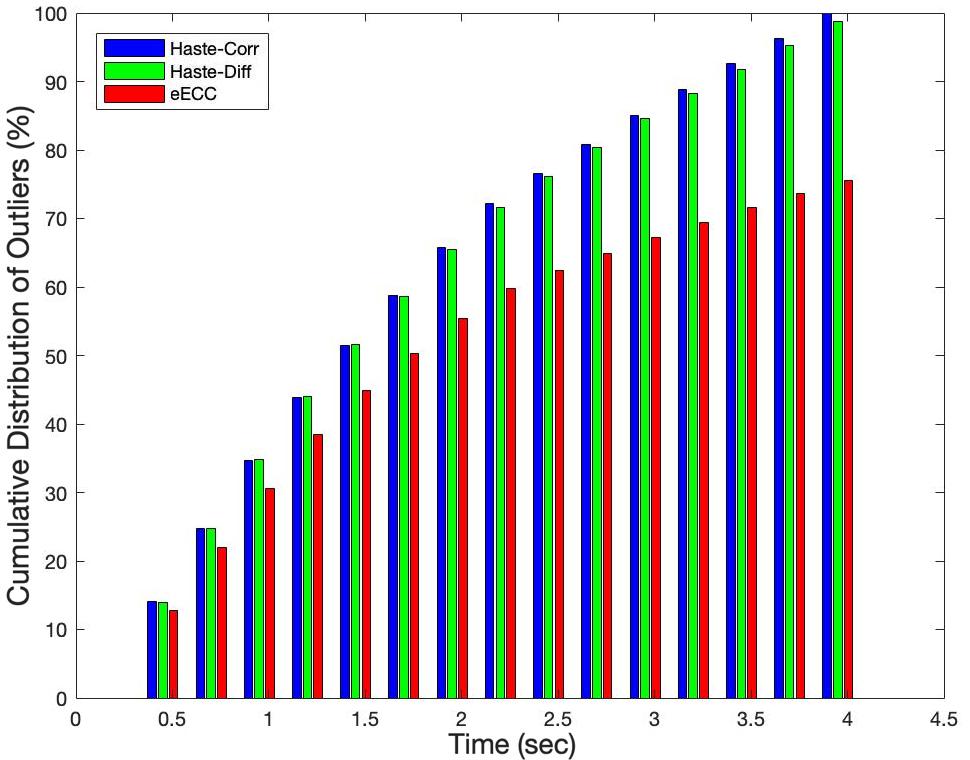} & 
        \includegraphics[width = 0.30\textwidth]{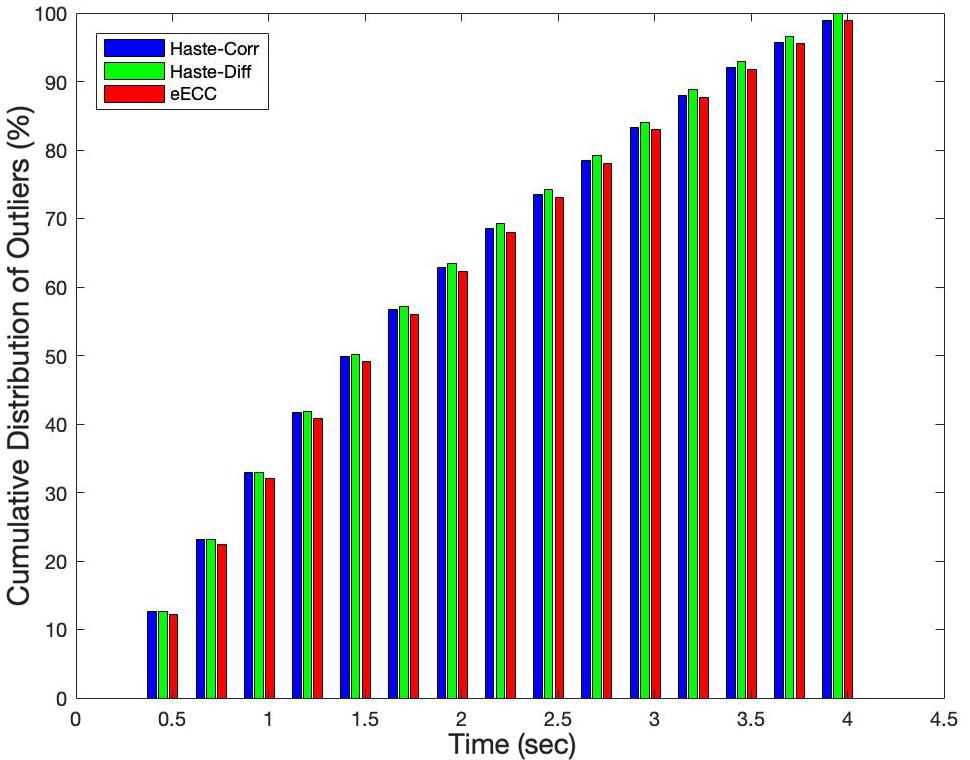} \\
    \end{tabular}
    \caption{Reprojection error and the cumulative distribution of outliers for the "wallposter" scene}
    \label{fig:mesh2}
\end{figure}
Fig.\ref{fig:mesh1}.(a) shows trajectories of tracked seeds on the image plane.
As mentioned, Haste baselines discretize the parameter space and analytically search for the optimum update. This technique results in quantized trajectories with state jumps. Instead, eECC provides continuous trajectories with very smooth transition between states. Since the algorithms build templates from motion-corrected events, the higher the tracking accuracy is, the sharper the template content is. Fig.\ref{fig:mesh1}.(b) shows templates that regard the star shape of Fig.\ref{fig:mesh1}.(a), obtained from the algorithms for the same seed and lifetime. eECC builds a sharper density map with a smaller active area that generates events. Recall that a perfect tracker would provide a quite small active area, that is, the density map would be similar to an edge map. 
\begin{figure}
\centering
     \begin{tabular}{c c c}
      Boxes translation & Boxes rotation & Boxes 6dof \\
  \includegraphics[width =0.30\textwidth]{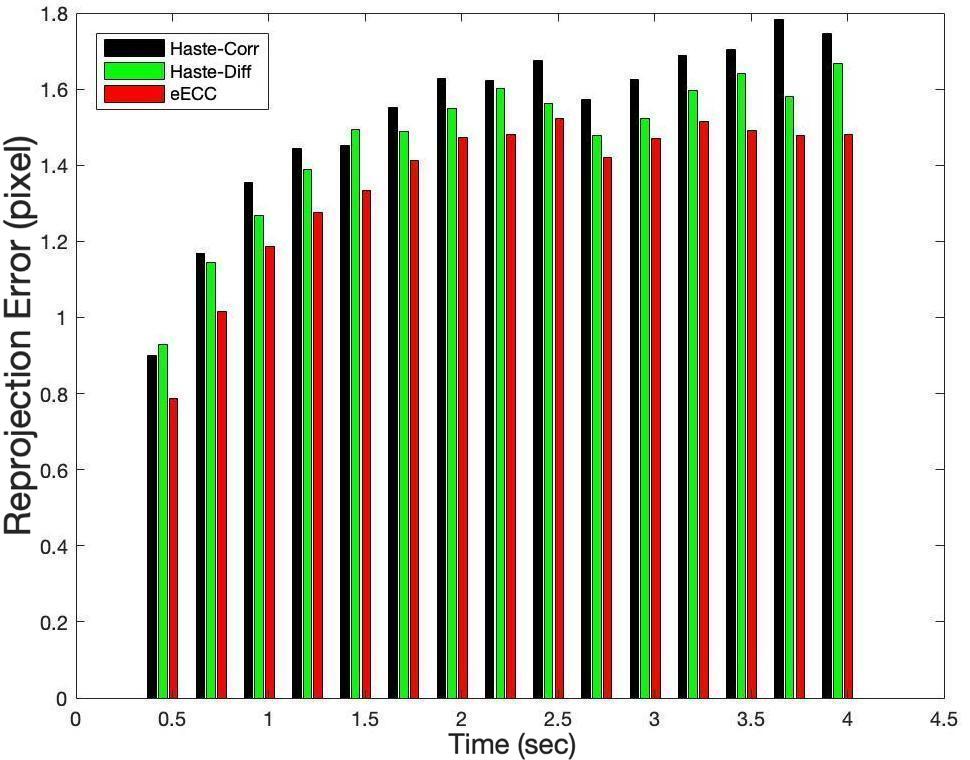}  & 
  \includegraphics[width = 0.30\textwidth]{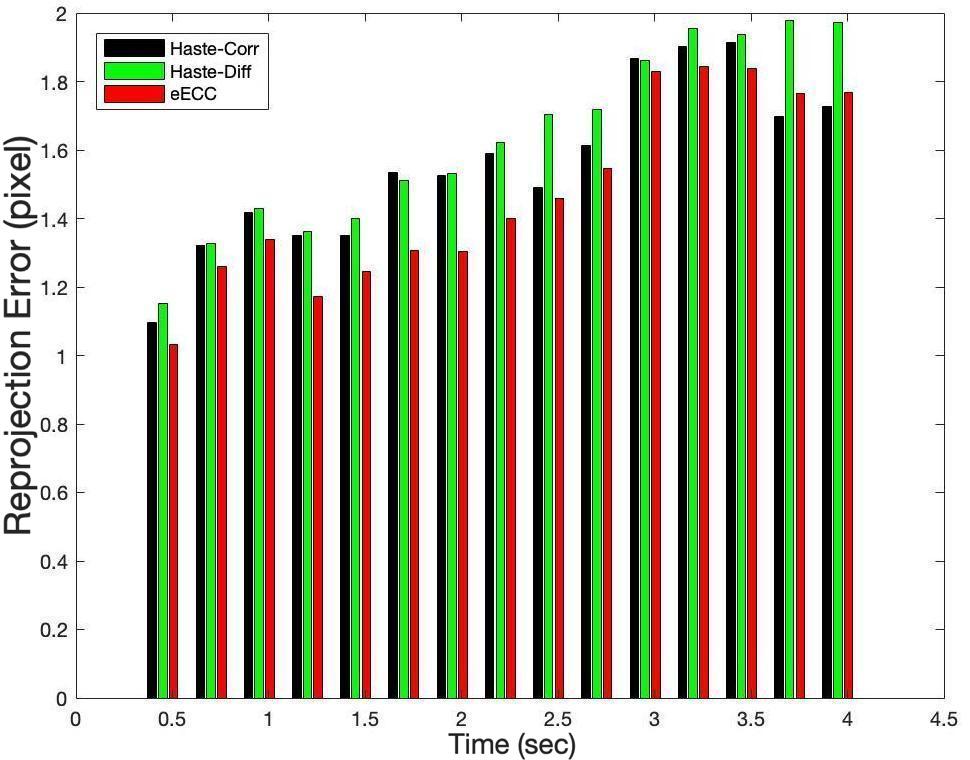}  & 
   \includegraphics[width = 0.30\textwidth]{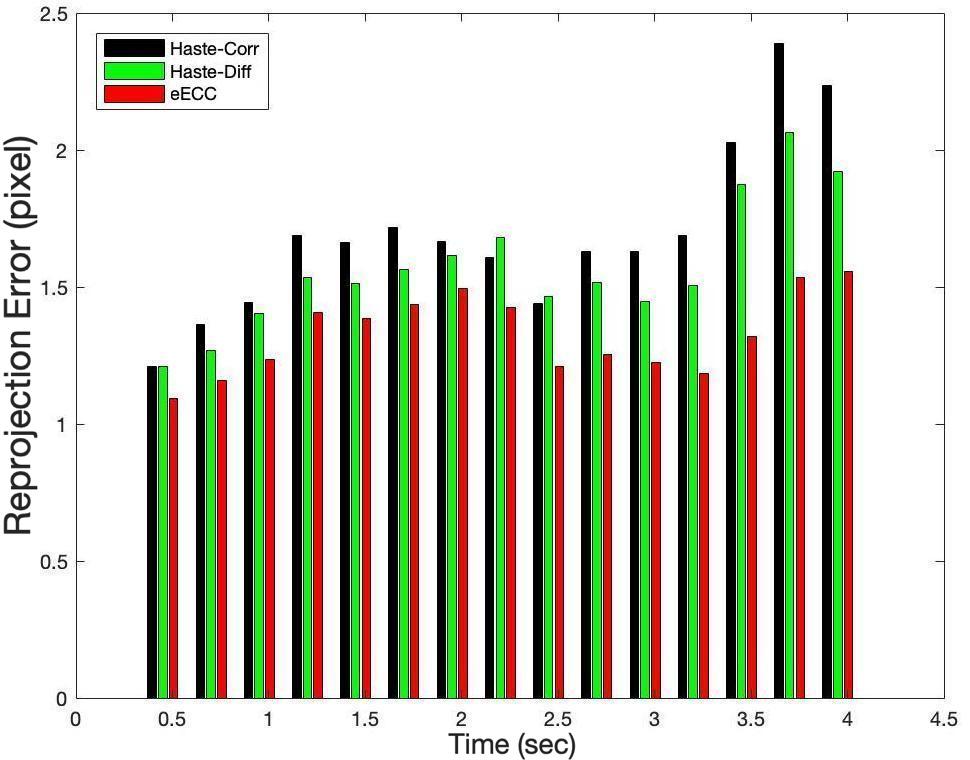}\\
  \includegraphics[width = 0.30\textwidth]{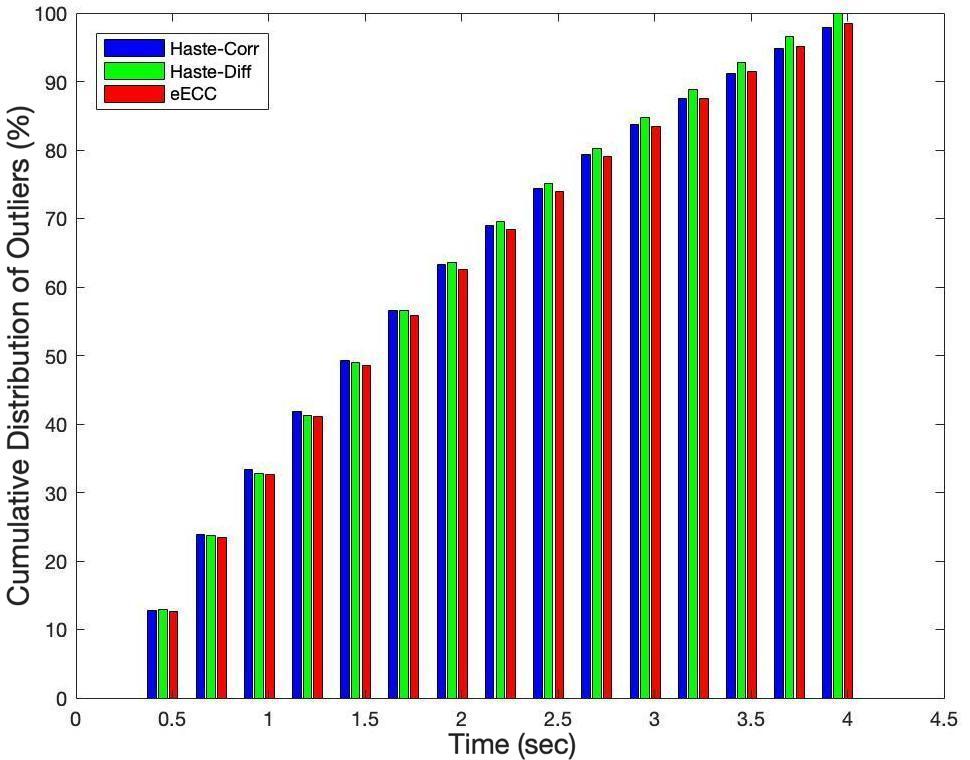} &
  \includegraphics[width = 0.30\textwidth]{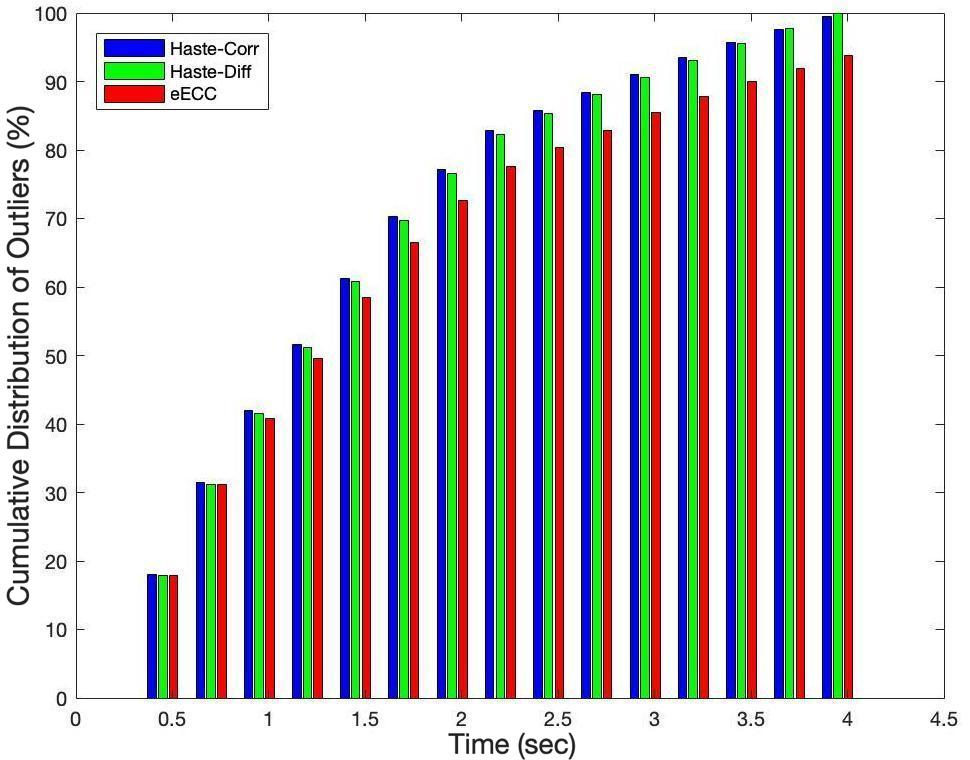} & 
        \includegraphics[width = 0.30\textwidth]{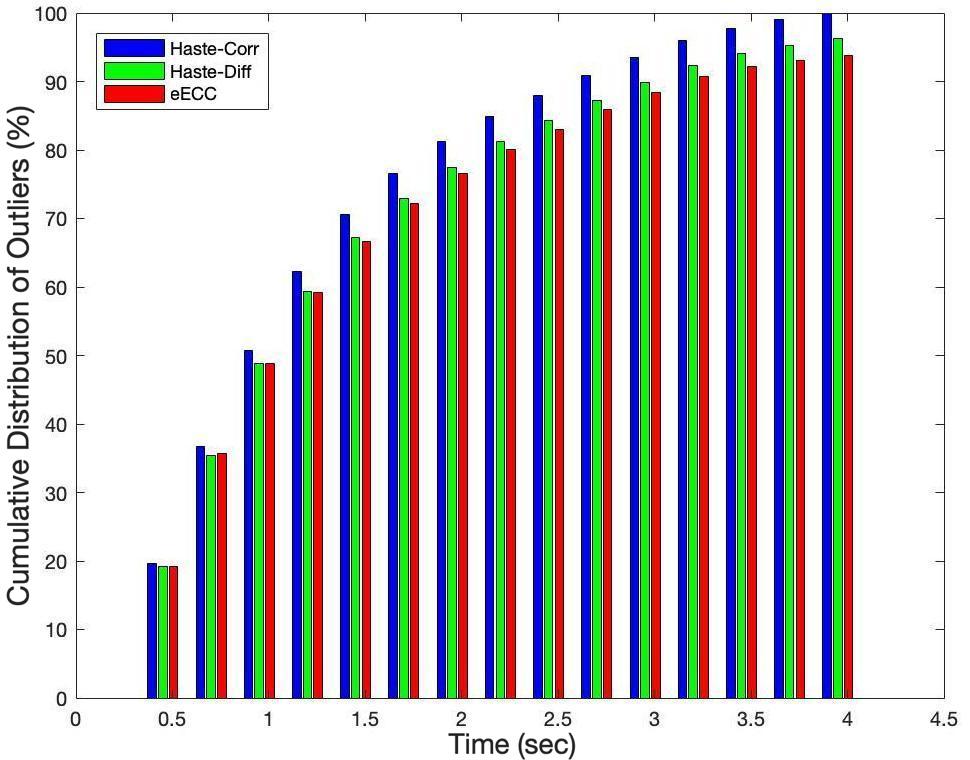} 
    \end{tabular} 
    \caption{Reprojection error and the cumulative distribution of outliers for the "boxes" scene}
    \label{fig:mesh3}
\end{figure}

\begin{figure}
\centering
\begin{tabular}{c c c}
     Dynamic translation & Dynamic rotation & Dynamic 6dof\\
  \includegraphics[width =0.30\textwidth]{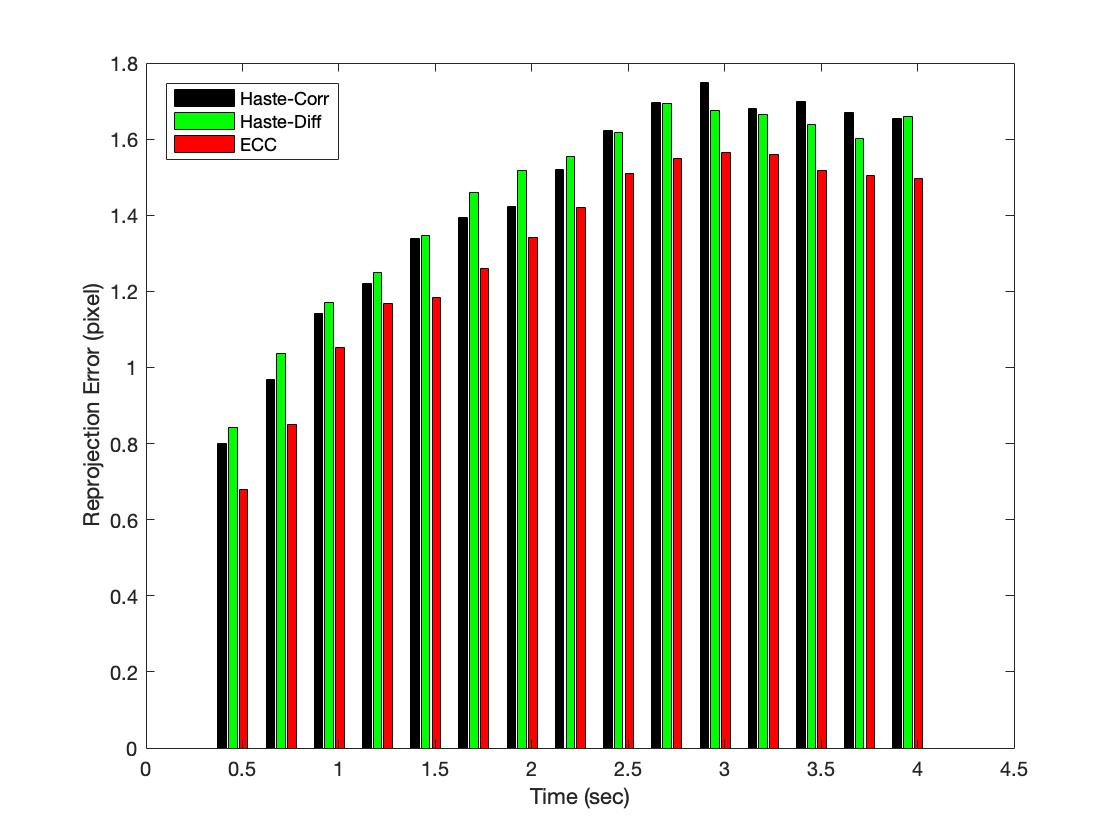}  &
  \includegraphics[width = 0.30\textwidth]{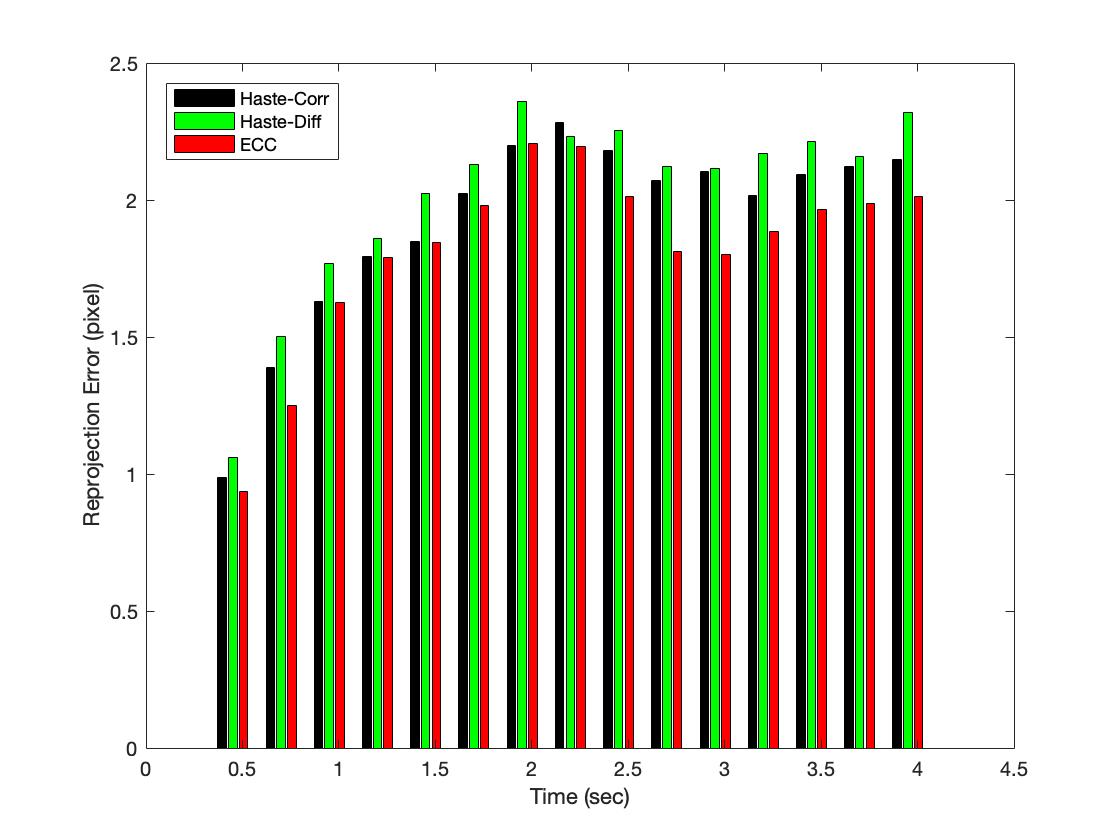}  & 
  \includegraphics[width = 0.30\textwidth]{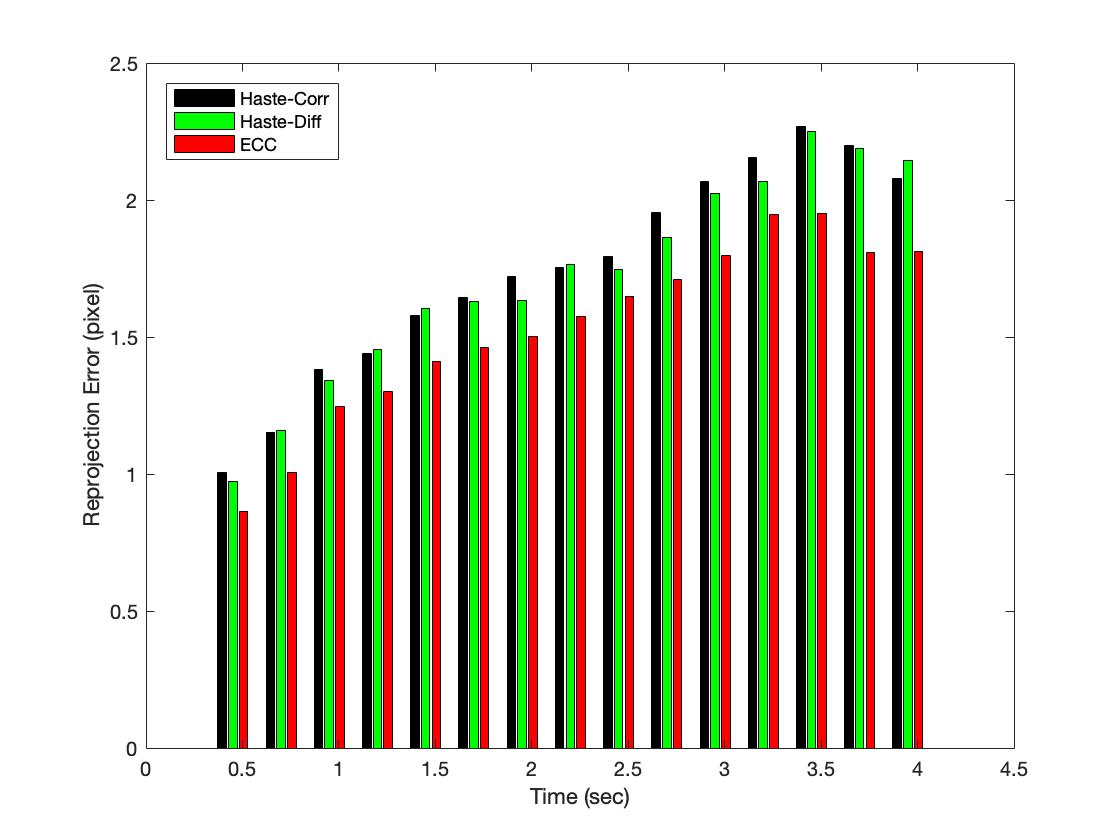}\\
  
  \includegraphics[width = 0.30\textwidth]{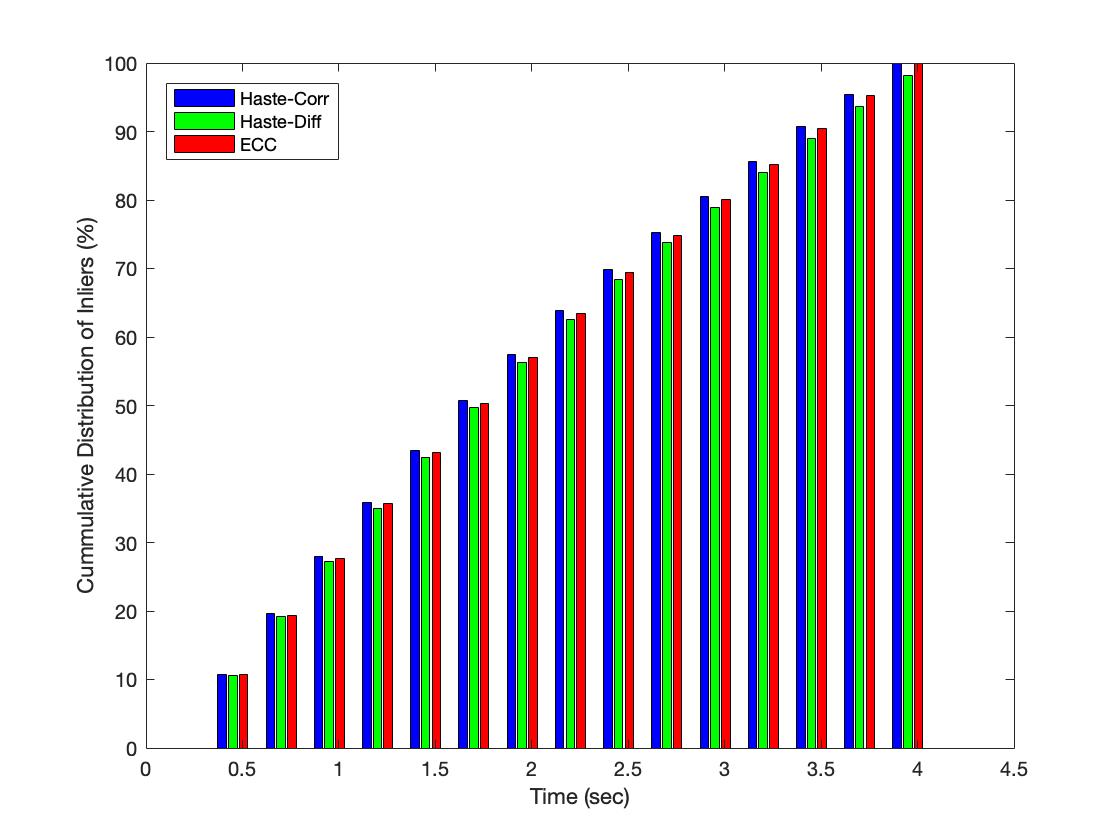} 
 &
 \includegraphics[width = 0.30\textwidth]{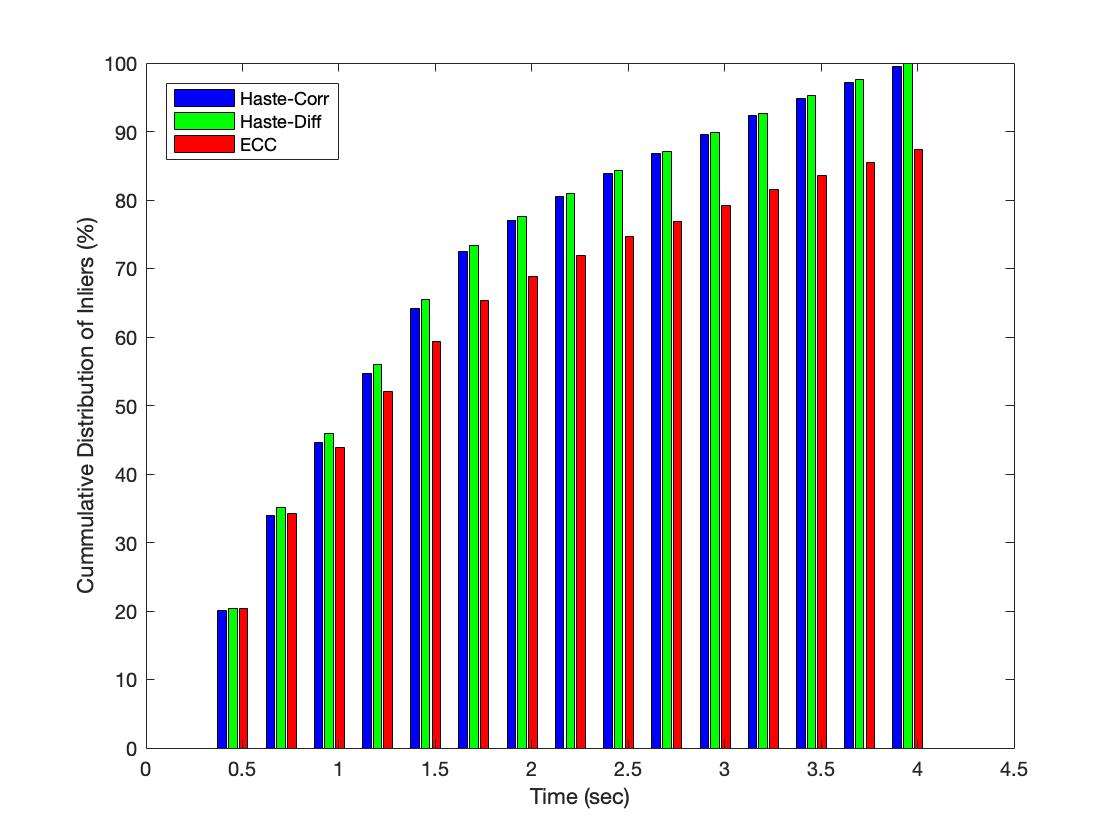} & 
        \includegraphics[width = 0.30\textwidth]{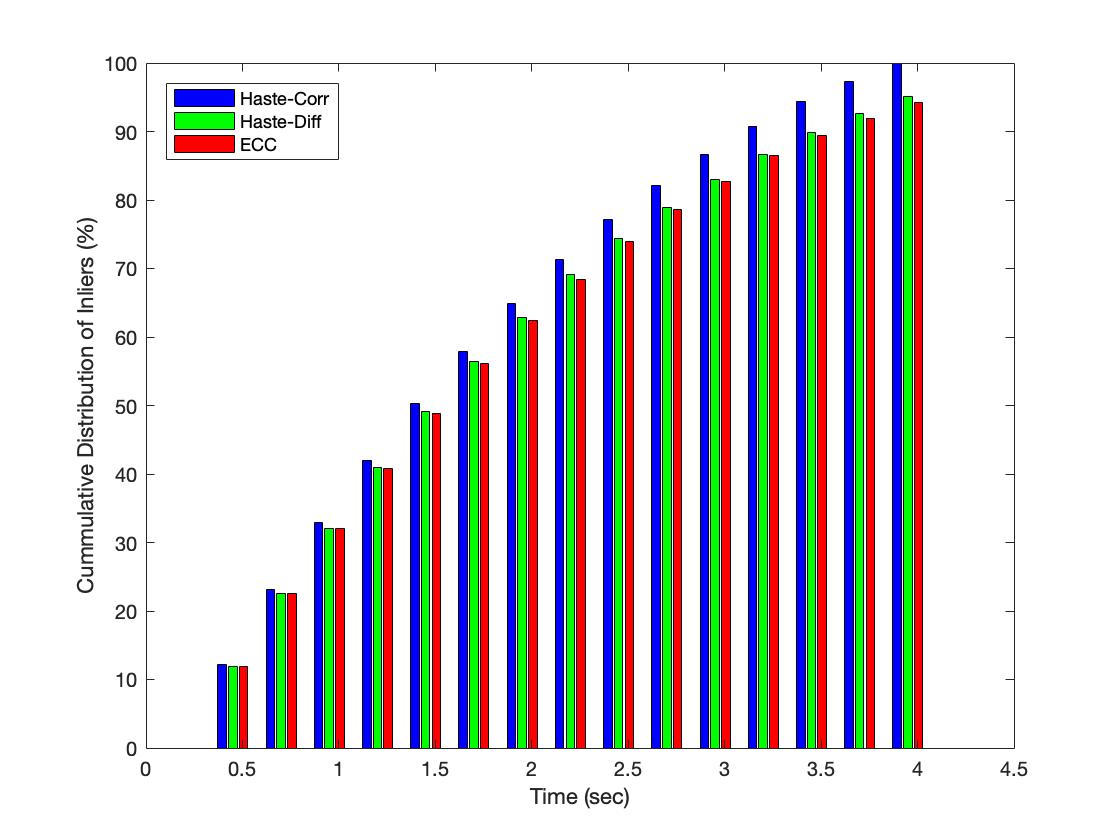} \\
    \end{tabular} 
\caption{Reprojection error and the cumulative distribution of outliers for the "dynamic" scene} \label{fig:mesh4}
\end{figure}

\begin{figure}
    \centering
    \begin{tabular}{c c c}
HDR poster & HDR boxes\\
  \includegraphics[width = 0.32\textwidth]{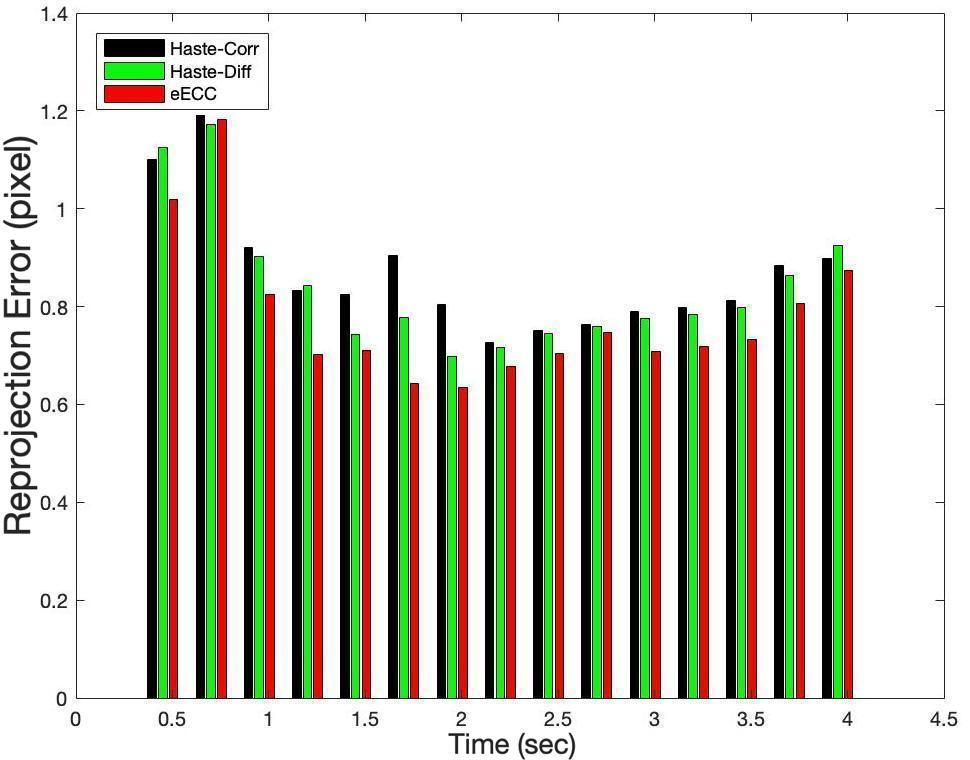}   &  \includegraphics[width = 0.32\textwidth]{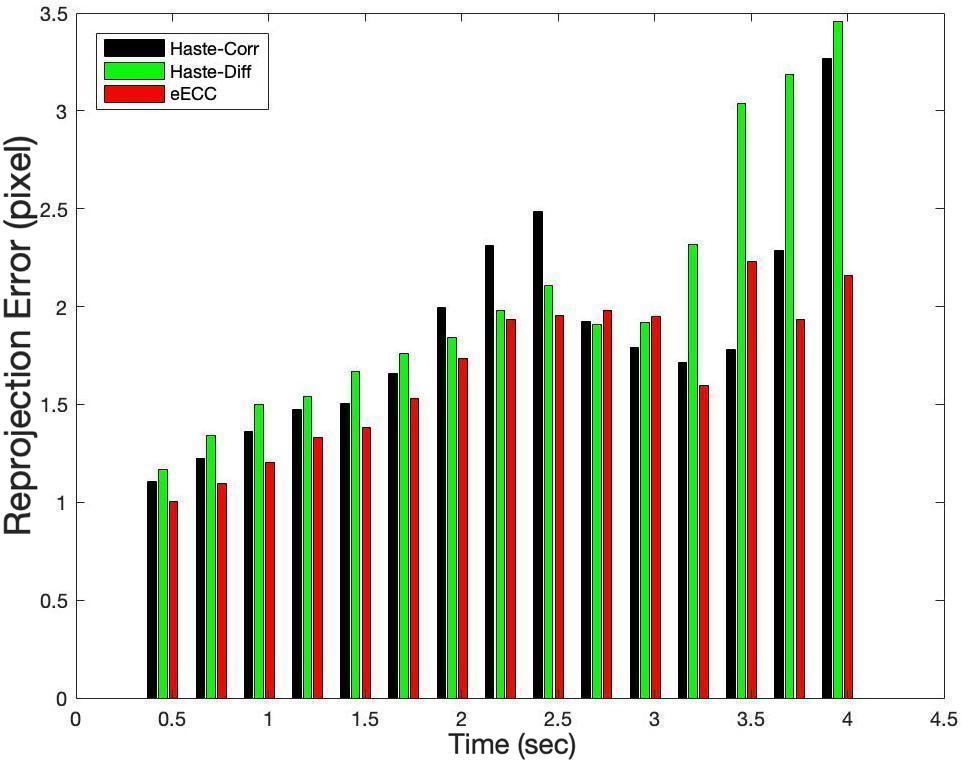}
\\
  
  \includegraphics[width = 0.32\textwidth]{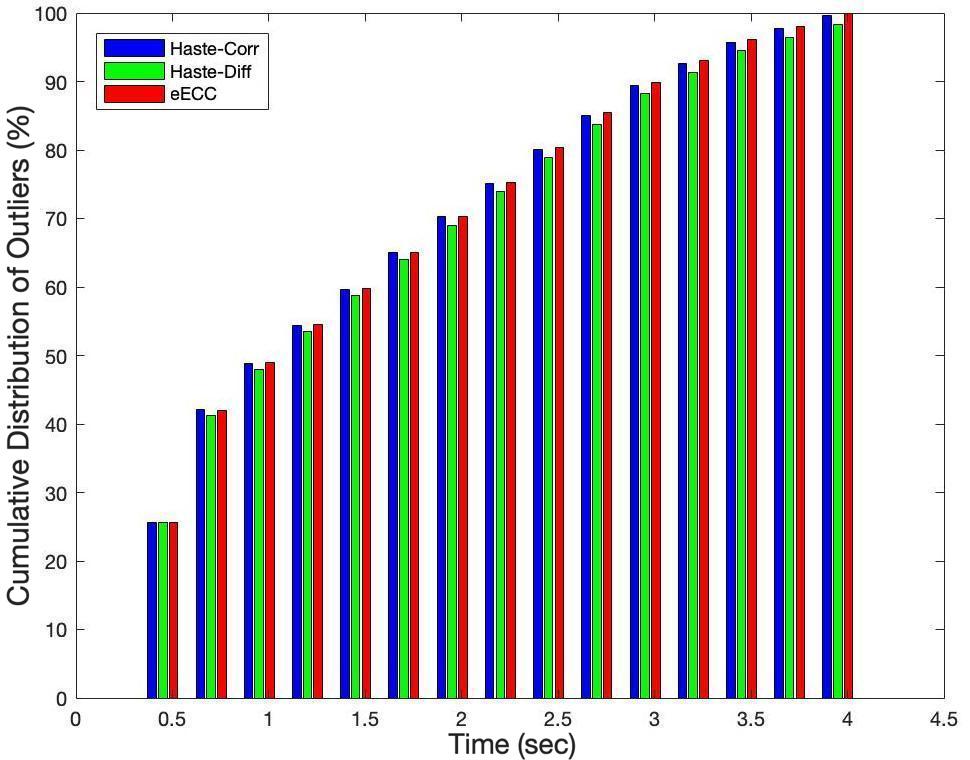} &
        \includegraphics[width = 0.32\textwidth]{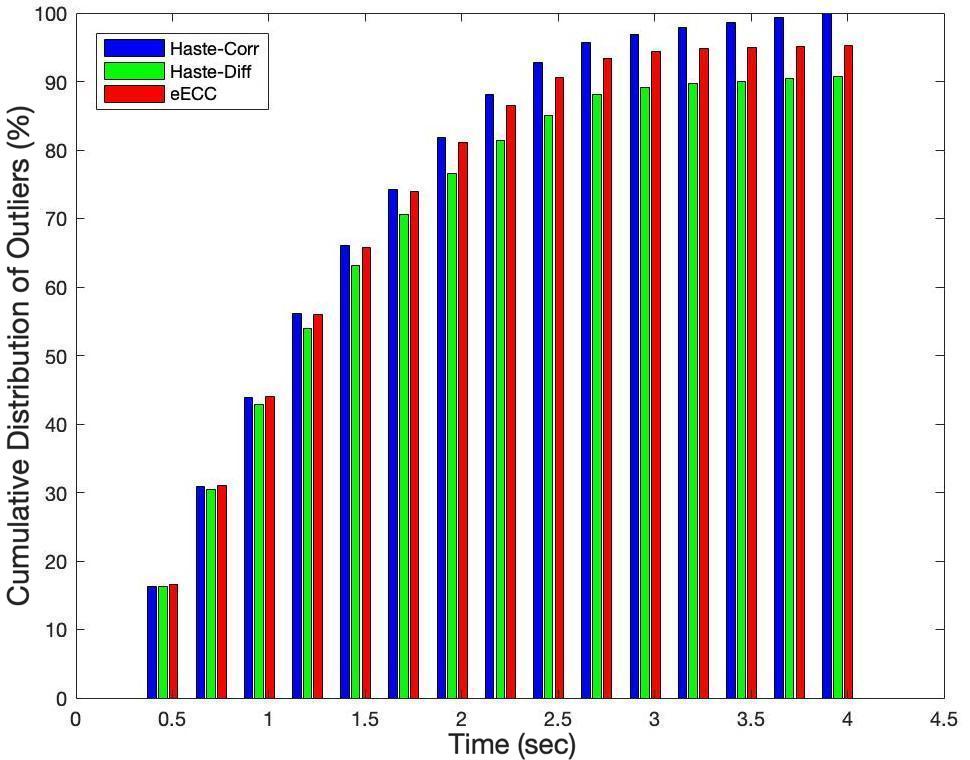} 
    \end{tabular}
    \caption{Reprojection error and the cumulative distribution of outliers for HDR recordings}
    \label{fig:mesh5}
\end{figure}

In Figs. \ref{fig:mesh2}-\ref{fig:mesh5}, the obtained reprojection errors and the cumulative distributions of outliers 
across four (4) different scenarios are shown. 
Under the first scenario, a 2D scene (wallposter) is recorded from a camera that undergoes translational, rotational and 6DoF motion respectively with increasing speed; under the second scenario, a highly textured 3D scene (boxes) is recorded from a camera that undergoes similar motions; under the third scenario, a 3D scene (Office with moving person) is recorded from a camera that undergoes similar motions; finally, under the fourth scenario, two scenes ( a textured flat wallposter and boxes) are captured in high illumination conditions.
eECC outperforms Haste baselines w.r.t. accuracy since it keeps the reprojection error at lower levels, being the maximum error difference equal to 1.5 pixel in "boxes" scene with 6DoF motion. As far as the feature age is concerned, eECC attains similar or lower percentages of lost tracks.

It is important to recall the low resolution of the dataset, that is, the decrease of reprojection error for the more-meaningful long tracks ($>2sec$), averaged over all the datasets, is  larger than $0.2$ pixel at $240\times180$ resolution. Such a decrease is mapped to $0.2/0.375 \simeq 0.53$  pixels at VGA resolution, which is typically used by real-time SLAM systems. Reducing the reprojection error by half pixel at VGA resolution is an improvement with strong impact on tracking the pose of the body.
 
A fair run-time and complexity comparison is not easy because of continuous nature of eECC and the discrete parameter space of Haste. The approach of Haste has the advantage of low complexity when a "regular event" is processed, that is, when the algorithm decides not to change the state. However, the balance between regular and state events depends on i) how detailed the discretization of the parameter space is and ii) the underlying parametric motion model. As in ~\cite{alzugaray_BMVC20_haste}, we here used the default $11$ hypotheses of Haste for the euclidean motion model ($3$ parameters). A smaller quantization step or an affine motion model ($6$ parameters) would exponentially increase the number of hypotheses, and in turn the probability to process a state event. E.g.,~\cite{alzugaray_3DV19_haste} considers $0.5$ pixel while it suggests that more complex perturbations could be employed (\cite{alzugaray_3DV19_haste} and \cite{alzugaray_BMVC20_haste} consider either pure translational or pure rotational perturbations).
Again, if one considers the low resolution and the mapping to VGA, $1$ pixel or $4$ degrees may be seen as a large perturbations (the tranlation corresponds to $\sim2.5$ pixel offset at VGA resolution) and the probability to vote for a state event is quite low. Whenever a state event is processed, Haste needs to reinitialize the hypotheses. Instead, the time complexity of eECC does not depend on these parameters.

\begin{table} 
\centering
\begin{tabular}{|l||c|} 
\hline
Tracker & State Event Time [$\mu$s/ev] \\ 
\hline  \hline
HasteDifference$^\ast$ & 31.8\\
\hline
HasteCorrelation$^\ast$ & \textbf{18.8}\\
\hline
\hline
eECC (non-incremental) & 43.2\\
\hline
$\text{eECC}$ & \textbf{11.8} \\
\hline
\end{tabular}
\vspace { 5 pt}
\caption{Computational time of trackers per (state) event}
\label{time_table}
\end{table}
In Table \ref{time_table}, we report the processing time per event on a recent Core-i5 (1GHz), including the times of the non-incremental version and the times of Haste methods for state events. Although eECC process (state) events $40\%$ faster than Haste Correlation*, the total tracking time of Haste baselines is lower because $\sim99\%$ of events turn out to be regular events for the specific motion model and discretization (1 pixel translation or 4 degree rotation). 

A similar strategy of skipping some events based on some criteria could be adopted by eECC, e.g. updating the quantities in Eq. (\ref{equs}) only when the correlation coefficient exceeds a predefined threshold. In this paper, we introduce the algorithm and primarily investigate the tracking accuracy and the potential of eECC to keep tracks alive; we leave the efficient approximations for a future work.

\section{Conclusions}\label{sec:conclusions}
In this paper, we propose a novel direct matching scheme for asynchronous event tracking. Based on the enhanced correlation coefficient (ECC) criterion, we formulate a lightweight version, called eECC, that process at every step the minimal information of a single event. Our experimental results show that eECC achieves state-of-the-art results in terms of tracking accuracy and feature age, in the context of asynchronous event-based tracking. As a future work, we intend to develop extensions that further reduce the complexity without compromising the accuracy, while increasing the feature lifetime.


\bibliographystyle{IEEEtran}
\bibliography{main}
\typeout{get arXiv to do 4 passes: Label(s) may have changed. Rerun}
\end{document}